\documentclass[preprint,12pt]{elsarticle}

\usepackage{lineno,hyperref}
\modulolinenumbers[5]

\journal{Journal of \LaTeX\ Templates}
\usepackage{mathrsfs} 
\usepackage{amsmath}
\usepackage{color}
\usepackage{amssymb}
\usepackage[ruled]{algorithm2e}
\usepackage{algpseudocode}
\usepackage{amsmath}
\usepackage{graphics}
\usepackage{epsfig}
\usepackage{graphicx}  %Required
\usepackage{mathrsfs} 
\usepackage{soul} 
\usepackage{multirow}
\usepackage{booktabs}
\usepackage{soul}
\usepackage{xcolor}

\usepackage{subfigure} 

\date{}

\begin{document}

\begin{frontmatter}

\title{Brain-inspired and Self-based Artificial Intelligence}

\author[1,2,4,3,5,8]{Yi Zeng\corref{cor1} \fnref{fn1,fn2}}
\author[1,2,4,8]{Feifei Zhao \fnref{fn1}}
\author[1]{Yuxuan Zhao \fnref{fn1}}
\author[2,4,8]{Dongcheng Zhao \fnref{fn1}}
\author[2,4,8]{Enmeng Lu\fnref{fn1}}
\author[1,2,4,5,8]{Qian Zhang}
\author[1,8]{Yuwei Wang}
\author[1,5]{Hui Feng}
\author[1,5]{Zhuoya Zhao}
\author[1,5]{Jihang Wang}
\author[1,5]{Qingqun Kong}
\author[1,2,4,8]{Yinqian Sun}
\author[1,5]{Yang Li}
\author[1,5]{Guobin Shen}
\author[1,5]{Bing Han}
\author[1,5]{Yiting Dong}
\author[1,5]{Wenxuan Pan}
\author[1,5]{Xiang He}
\author[6,1,8]{Aorigele Bao}
\author[6,1]{Jin Wang}

\fntext[fn1]{These authors contributed equally.}
\fntext[fn2]{Lead contact}
\cortext[cor1]{Correspondence: yi.zeng@braincog.ai, yi.zeng@ia.ac.cn}

\address[1]{Brain-inspired Cognitive AI Lab, Institute of Automation, Chinese Academy of Sciences, China.}
\address[2]{Beijing Key Laboratory of Safe AI and Superalignment, China.}
\address[4]{Beijing Institute of AI Safety and Governance, China.}
\address[3]{Key Laboratory of Brain Cognition and Brain-inspired Intelligence Technology, Chinese Academy of Sciences, China.}
\address[5]{University of Chinese Academy of Sciences, China.}
\address[6]{Department of Philosophy, School of Humanities, University of Chinese Academy of Sciences, China.}
\address[8]{Long-term AI, China.}

\end{frontmatter}

%\begin{affiliations}
% \item Institute of Automation, Chinese Academy of Sciences.
%\end{affiliations}

\section*{Summary}

The question "Can machines think?" and the Turing Test to assess whether machines could achieve human-level intelligence is one of the roots of AI. With the philosophical argument "I think, therefore I am", this paper challenge the idea of a "thinking machine" supported by current AIs since there is no sense of self in them. Current artificial intelligence is only seemingly intelligent information processing and does not truly understand or be subjectively aware of oneself and perceive the world with the self as human intelligence does. In this paper, we introduce a \textbf{Br}ain-\textbf{i}nspired and \textbf{Se}lf-based \textbf{A}rtificial \textbf{I}ntelligence (BriSe AI) paradigm.
This BriSe AI paradigm is dedicated to coordinating various cognitive functions and learning strategies in a self-organized manner to build human-level AI models and robotic applications. Specifically, BriSe AI emphasizes the crucial role of the Self in shaping the future AI, rooted with a practical hierarchical Self framework, including Perception and Learning, Bodily Self, Autonomous Self, Social Self, and Conceptual Self. The hierarchical framework of the Self highlights self-based environment perception, self-bodily modeling, autonomous interaction with the environment, social interaction and collaboration with others, and even more abstract understanding of the Self. Furthermore, the positive mutual promotion and support among multiple levels of Self, as well as between Self and learning, enhance the BriSe AI's conscious understanding of information and flexible adaptation to complex environments, serving as a driving force propelling BriSe AI towards real Artificial General Intelligence.

\section{Introduction}

In his 1950 paper "Computing Machinery and Intelligence," Alan Turing asked the question "Can machines think?" and introduced the Turing Test to verify whether machines could achieve human-level intelligence~\cite{turing1950}. However, Artificial Intelligence (AI) that meets the external requirements of behaviorism is not necessarily "intelligent" in the true sense, as evidenced by John R. Searle's Chinese Room thought experiment~\cite{searle1986minds}. As René Descartes said, "I think, therefore I am," implying that being aware of one's thoughts is a crucial element of self-consciousness~\cite{descartes1987discours}. However, current AI is merely an appearance of intelligence through information processing, rather than being subjectively aware of oneself, perceive and understand the world from the self perspective and learn from others or perspectives of others like human intelligence. Without the construction and sense of self for AI, we argue there is fundamentally impossible to develop a machine that think in the real sense. Therefore, the concept of Self should play a crucial role in the development of AI.

Some research has already been exploring how to bestow AI and robots with self-awareness akin to humans. Tony J. Prescott argues that, considering the complexity of the Self knowledge, a unified cognitive architecture is indispensable for realizing a multi-faceted self-model~\cite{Prescott2020}. In fact, given that self-awareness is a vital component of human mind, AI researchers have, for decades, closely aligned their efforts with simulating human minds through the construction of cognitive architectures. Cognitive architectures simulate and understand the human mind through computational modeling, e.g., memory, learning, reasoning, and other cognitive mechanisms and processes. Allen Newell champions that human cognition and mind function as a unified cognitive architecture~\cite{newell1994unified}. Inspired by this, John R. Anderson proposed the Adaptive Control of Thought-Rational (ACT-R) architecture~\cite{anderson2013language}, a cognitive psychology-based model that integrates the human associative memory model with generative systems architecture, covering perception, memory, decision-making, language, and motor cognitive functions. Allen Newell poured his proposed unified theories of cognition into the Soar cognitive architecture~\cite{laird1987soar}, which was then continuously developed by John E. Laird~\cite{laird2019soar}. Soar is committed to providing the underlying infrastructure that enables the system to perform a variety of cognitive tasks, capable of implementation of a general intelligence. Soar consists of several interacting task-independent modules: short-term and long-term memories, processing modules, learning mechanisms, and interfaces between them.

Unlike the top-down cognitive architecture design of ACT-R, Local, Error-driven and Associative, Biologically Realistic Algorithm (Leabra)~\cite{o1996leabra} begins from the basic neural mechanisms and integrates them into a unified framework from the bottom to up. Specifically, Leabra incorporates point neuronal computations, and different learning mechanisms organized to correspond to biologically plausible formulations. Then, neurons connect and interact in networks to form small-scale and large-scale brain regional structures. Various brain regional architectures organized and interacted with each other to perform different cognitive functions. SAL~\cite{ jilk2008sal} synthesizes two cognitive architectures, ACT-R and Leabra, treating them as modular and replaceable components, proposing a pluralistic approach reorganized to meet the needs of cognition. Connectionist Learning with Adaptive Rule Induction On-line (CLARION) ~\cite{sun2003tutorial} considers the distinction between implicit and explicit cognitive processes and captures their interactions integratively. In addition, Non-Axiomatic Reasoning System (NARS)~\cite{ wang1995non} shows an experience-based reasoning system using a formal term-oriented language. Semantic Pointer Architecture Unified Network (SPAUN)~\cite{eliasmith2012large} has a large-scale spiking neurons organized into subsystems that resemble specific brain regions, capable of performing multiple tasks.

In general, the above cognitive architectures synergize different sub-modules and sub-systems into a unified architecture for accomplishing the full range of cognitive tasks. They approach and simulate human cognitive processes as closely as possible at different levels, covering brain regions such as the prefrontal cortex, basal ganglia, posterior cortex, and hippocampus, performing cognitive tasks such as episodic learning, semantic learning, reinforcement learning, working memory, long-term memory, perception, and action selection. There are also cognitive architectures related to consciousness, such as the Global Workspace Theory~\cite{baars1993cognitive}, which postulates that attention can transform unconscious processes or modules into conscious awareness. Building on this foundation, Learning Intelligent Distribution Agent (LIDA)~\cite{baars2009consciousness} focuses on the computational modeling of consciousness and higher-order cognitive control.

Obviously, the theory and development of cognitive architectures show the great inspiration for AI, however existing cognitive architectures often use symbolic rule systems or neural networks for psychological computational modeling. Seamlessly leveraging cognitive architectures to develop advanced general-purpose AI still has room for multi-dimensional enhancements, such as human-level higher-order cognition (e.g., self), and self-organized synergistic cognitive functions to solve more complex cognitive tasks.

In this paper, we propose a \textbf{Br}ain-\textbf{i}nspired and \textbf{Se}lf-based \textbf{A}rtificial \textbf{I}ntelligence paradigm, namely BriSe AI, that commits to endowing AI with genuine understanding of self, environment, and others under self-supervision, facilitating a harmonious coexistence between humans and machines. BriSe AI integrates a feasible and practical framework of Self, including five levels of Self: Perception and Learning, Bodily Self, Autonomous Self, Social Self, and Conceptual Self. 

\begin{itemize}
  \item  \textbf{Perception and Learning.} The ability of an intelligent agent to perceive and identify the external environment, providing perceptual information for a higher level of Self.
  \item  \textbf{Bodily Self.} 
Intelligent agents' perception and cognition of their bodies, bodily self-modeling e.g., kinematic modeling of the body, perceiving body position, and predicting movements.
  
  \item  \textbf{Autonomous Self.} Emphasizing that intelligent agents autonomously interact and explore their surrounding environment, accomplishing tasks while acquiring self-experience and self-causal awareness.
  
  \item  \textbf{Social Self.} Towards social interactions, intelligent agents can distinguish themselves from others, and understand others' mental states and emotions based on self-experiences, perceptions, and cognition. Social Self enables agents to cooperate better with others, helps them understand social norms, and leads to moral and ethical decision-making.

  \item  \textbf{Conceptual Self.} The highest level of the entire self-model involves an abstract and conceptual understanding of the self's identity, roles, values, and goals, integrating and interacting with different levels of the self.

\end{itemize}

\begin{figure}[!htbp]
\centering
\includegraphics[scale=0.45]{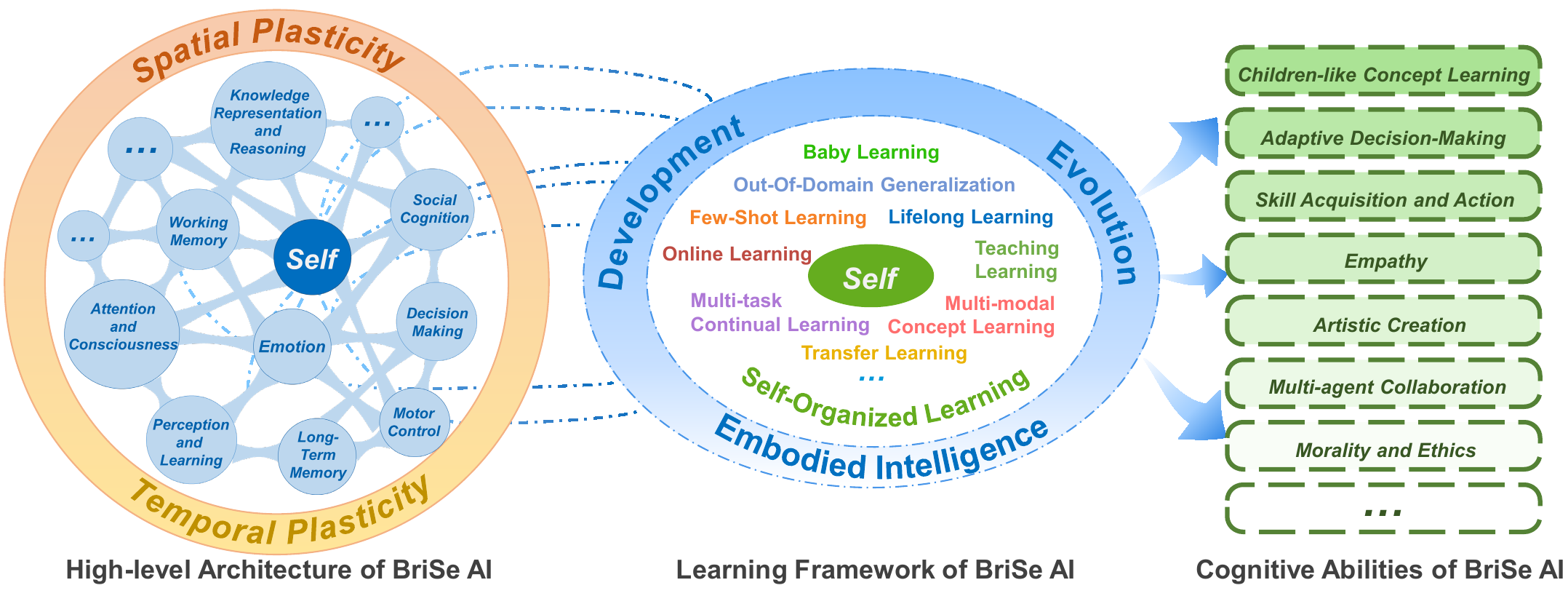}
\caption{The functional vision of BriSe AI.}
\label{BriSevision}
\end{figure}

This self-framework for AI is able to acquire increasingly complex cognitive abilities online and coordinate multiple cognitive functions and learning strategies in a self-organized manner. As shown in Figure~\ref{BriSevision}, the high-level architecture of BriSe AI uses the Self hierarchical framework to coordinate multiple cognitive functions such as perception and learning, working memory, long-term memory, decision making, motor control, knowledge representation and reasoning, emotion, attention and consciousness, social cognition into a self-organizing unified framework. Based on this, BriSe AI integrates various learning strategies including unsupervised learning, supervised learning, reinforcement learning, few-shot learning, continual learning, transfer learning, and multi-modal learning, etc. to achieve high-level cognition such as child-like concept learning, adaptive decision-making, skill acquisition and action, empathy, artistic creation, multi-agent collaboration, morality and ethics. Conversely, a higher-level Self forms a top-down feedback supervision, prompting fundamental learning strategies to leap from information processing to self-modulated information understanding.

To implement BriSe AI, we use an Spiking Neural Network (SNN)~\cite{maass1997networks} based brain-inspired cognitive intelligence engine (BrainCog)~\cite{zeng2023braincog}, seamlessly integrating BrainCog's components and cognitive functions to build advanced AI models and applications. The basic components of BrainCog include spiking neurons with different levels of granularity, biological learning rules, multiple encoding strategies, and various functional brain region models. Based on these fundamental components, BrainCog implements various learning strategies of spiking neural networks, such as unsupervised learning, supervised learning, associative learning, and reinforcement learning. This supports cognitive functions such as perception and learning, decision-making, motor control, knowledge representation and reasoning, social cognition, etc.

In addition to multi-scale spatial neuroplasticity principles at micro, meso, and macro scales, BriSe AI further integrates learning, developmental and evolutionary plasticity at different timescales. It also extends learning strategies such as transfer learning, continual learning, child-like learning, lifelong learning, multi-modal conceptual learning, and out-of-domain generalization. More importantly, BriSe AI realizes several cognitive capabilities around the Self such as self-world distinction, bodily self-perception, self-others distinction, cognitive empathy, and affective empathy. Through the self-organized collaboration of multiple learning strategies and cognitive abilities, as well as the mutual promotion between Self and learning, and among different levels of Self, BriSe AI is able to better understand self, environment and others, exhibit safe, moral behavior, and progress towards advenced 
general-purpose AI. 

\section{Hierarchical Framework of the Self for AI}

The study of self-concept and theories is primarily concentrated in the fields of philosophy and cognitive psychology, while research in the field of artificial intelligence is relatively limited and has not yet achieved complete unity. In 1988, Ulric Neisser proposed five kinds of self-knowledge~\cite{neisser1988five}, including ecological self, interpersonal self, temporally extended self, private self, and conceptual self. These several 'self' possess the following characteristics: 1) Ecological self has a point of view and a sense of body ownership. 2) Interpersonal self perceives others as agents similar to oneself and having empathy for them. 3) Temporally extended self is aware of your personal past and future. 4) Private self experiences a stream of consciousness and is aware of your inner life. 5) Conceptual self has a life story, personal goals, motivations, and values. These five kinds of self-knowledge proposed by Ulric Neisser are categorized and generalized from the perspective of the self that humans possess. Building upon this, Tony J. Prescott proposed Robot Self~\cite{Prescott2020}, which includes the situated self, agential self, spatiotemporal self, interpersonal self, conceptual self, and private self. 
The main difference from Ulric Neisser's five kinds of self is the addition of the agential self, emphasizing the characteristics of a robot actively seeking information, selecting actions that generate integrated behavior, and knowing what events it caused in the world. 

Overall, the existing two frameworks of Self are conceptually coarse-grained inductions of 'self', which are not feasible enough to implement in AI. The Self we endow for robots takes inspiration from the human self, and hierarchically and progressively enhances the AI's self level through self-organized coordination of brain-inspired cognitive functions and learning strategies. More importantly, it is essential to utilize the higher Self to supervise and modulate the fundamental cognitive functions to achieve a true understanding, and the robot's embodied ontology in conjunction with the application of artificial general intelligence.

\begin{figure}[!htbp]
\centering
\includegraphics[scale=0.47]{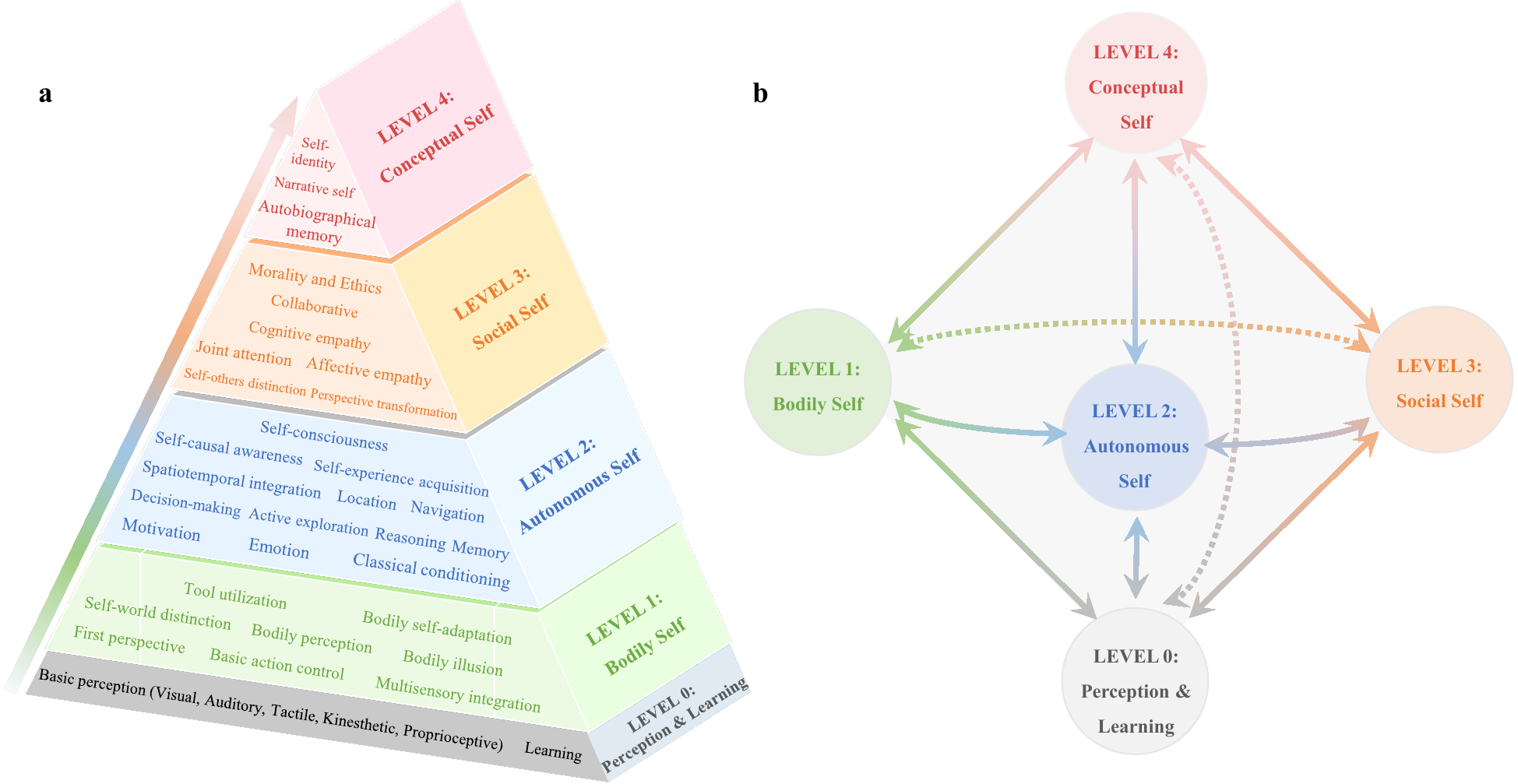}
\caption{The architecture of the Self for BriSe AI. \textbf{a.} The hierarchical framework of the Self for BriSe AI. \textbf{b.} The interaction between different levels of Self.}
\label{self}
\end{figure}

In this paper, we propose a hierarchical framework of Self to support BriSe AI. Our hierarchical self-architecture provides a feasible and practical framework for self-modeling of intelligent agents, enabling them to better understand and adapt to themselves at different levels, and offering new insights and directions for the study of self-awareness in the field of artificial intelligence. As shown in Figure~\ref{self}a, our proposed hierarchical self-framework contains five levels, each corresponding to a different stage of self-awareness and understanding of the intelligent agent.

\begin{enumerate}
    \item LEVEL 0: Perception \& Learning encompasses the ability of intelligent agents to perceive the external environment and fundamental learning. At this level, intelligent agents establish preliminary self-awareness through perceiving and identifying external information.
    \item LEVEL 1: Bodily Self focuses on the intelligent agent's awareness of its own body. At this level, the agent gradually develops a model for bodily kinematics, positional awareness, and motion prediction, laying the foundation for higher-level self-awareness.
    \item LEVEL 2: Autonomous Self emphasizes the autonomy and initiative of intelligent agents. At this level, agents through active exploration of the environment and task execution, achieve the acquisition of self-experiences, providing support for more complex cognitive tasks.
    \item LEVEL 3: Social Self introduces the social dimension, focusing on the role and relationships of intelligent agents in social interactions. At this level, agents simulate the thoughts, emotions, and intentions of others, achieving perspective-taking and empathy, thereby deepening their understanding of themselves in social interactions.
    \item  LEVEL 4: Conceptual Self encompasses an abstract understanding of the agent's identity, values, and goals. By connecting individual experiences with conceptual frameworks, intelligent agents can engage in higher-order thinking, planning, and decision-making, demonstrating higher levels of intelligence in complex cognitive tasks.
\end{enumerate}

The different levels of the Self are interconnected and interact collaboratively, as shown in Figure ~\ref{self}b. Perception \& learning, as the foundation of self-formation, assist intelligent agents in gradually acquiring the bodily self, autonomous self, social self, and conceptual self. As the Self's level increases, the higher Self provides top-down feedback to the lower Self, monitoring and scrutinizing to enhance fundamental cognitive and understanding skills. As a result, the mutual influence and support among different levels of Self endow BriSe AI with conscious, human-level thinking and understanding abilities, enabling it to flexibly adapt to complex environments.

\subsection{Perception and Learning} 
At the L0 level, intelligent agents perceive and recognize environmental information through embedded sensors, involving data input from multiple sensors such as vision, audition, touch, kinesthetic, and proprioceptive, along with underlying pattern recognition processes. This enables intelligent agents to perceive the characteristics of their surroundings and realize basic learning and information processing capabilities. With the top-down modulation by the higher-level self, conscious perception and learning will make the leap from information processing to information understanding.

\subsection{Bodily Self} 
At the L1 level, an intelligent agent undergoes a complex process of constructing a body model through perceiving information and random movements. This process involves modeling the kinematics of the body, perceiving body positions, and predicting movements. In this early stage, the intelligent agent gradually forms awareness and cognition of its own body state, laying a solid foundation for subsequent levels of self-modeling. Specifically, the L1 level includes several elements at lower levels. The first perspective allows the intelligent agent to perceive the environment from its own perspective, basic action control enables it to execute basic movements and actions, and multisensory integration allows it to integrate various sensory information. These elements collectively constitute the initial cognition of the intelligent agent about its own body, enabling it to perceive its position, movements, and basic actions in the environment.

On this foundation, the intelligent agent gradually achieves intermediate-level self-modeling. Self-world distinction enables the agent to have a clearer understanding of its relationship with the external world, Bodily perception allows for a deeper perception of various aspects of the body, and bodily illusion adds complexity to the cognition of body states. These intermediate-level elements provide the intelligent agent with a more refined and complex body model.

Ultimately, the L1 level realizes advanced-level elements, including tool utilization and bodily self-adaptation. This indicates that the intelligent agent not only flexibly utilizes tools for interaction with the environment but also adapts to changes in its own body state, achieving higher-level self-awareness and self-adjustment capabilities. Therefore, the L1 level provides a crucial foundation for the development of intelligent agents in cognition and behavior, laying a solid cornerstone for their higher-level learning and adaptation abilities.

\subsection{Autonomous Self}
At the L2 level, an intelligent agent engages in active exploration of its environment, performs various tasks, and strives to acquire self-experiences. This process involves proactive perception of the surrounding environment, active execution of tasks, and the collaborative functioning of perception and decision-making systems. Through this collaboration, the intelligent agent gradually learns effective behaviors and strategies, accumulating richer experiences to establish a solid decision-making foundation for subsequent levels of self-modeling. The L2 level encompasses several lower-level elements, including motivation, decision-making, reasoning, and emotion. These fundamental elements work together to enable the intelligent agent to form initial cognition and experiences during active exploration and task execution.

Building upon this foundation, the L2 level incorporates more complex elements, such as classical conditioning, active exploration, and memory. The processes of active exploration and memory enable the intelligent agent to gather more information from the environment, storing and recalling past experiences, laying the groundwork for the development of higher-level cognition.
Further in the evolution of the L2 level, elements like spatiotemporal integration, location, and navigation emerge. These elements enhance the intelligent agent's ability to understand and navigate the spatial and temporal aspects of its environment, improving its performance in executing tasks within complex surroundings.

Finally, the L2 level endows the intelligent agent with the ability of self-causal awareness, where it becomes conscious of the consequences of its actions. Simultaneously, through active exploration and task execution, the intelligent agent acquires its own experiences, leading to self-experience acquisition. This culminates in the emergence of an intelligent agent possessing self-consciousness, signifying its awareness of its own existence and the impact of its actions in the environment. This achievement represents a higher level of self-awareness and consciousness, as the intelligent agent not only perceives and understands the external environment but also recognizes its own presence and influence within that environment.

\subsection{Social Self}
At the L3 level, intelligent agents engage in simulating the thoughts, emotions, and intentions of others to achieve perspective-taking and empathy. This advanced process involves reasoning about the mental states of others, understanding social norms, and adjusting their behavior by predicting the reactions of others. The social self-level plays a crucial role in the development of intelligent agents, helping them integrate better into social interactions and establish effective social relationships. At the L3 level, several lower-level elements are included, such as self-others distinction (distinguishing between oneself and others) and perspective transformation. These elements enable intelligent agents to have a clearer understanding of the differences between themselves and others in social interactions and to comprehend issues from different viewpoints.

On this basis, the L3 level incorporates more complex elements, including joint attention (sharing attention with others) and affective empathy. Joint attention allows intelligent agents to focus on specific aspects collaboratively, while affective empathy enables them to feel and understand the emotions of others, further enhancing adaptability to social scenarios. Further development in the L3 level includes elements like cognitive empathy and collaboration. This implies that intelligent agents can not only collaborate with others but also understand and share the cognitive experiences of others, promoting deeper levels of social interaction.

Eventually, the L3 level achieves morality and ethics, providing intelligent agents with the ability for moral judgment and ethical decision-making. This signifies that intelligent agents can consider moral standards and ethical principles in social interactions, offering higher-level guidance and norms for their participation in social activities. Therefore, the L3 level marks a profound understanding of social cognition and emotions in intelligent agents, enabling them to engage more effectively in social interactions and establish positive social relationships.

\subsection{Conceptual Self}

At the L4 level, intelligent agents develop conceptual self-awareness, involving an abstract understanding of their own identity, values, and goals. By connecting individual experiences with abstract conceptual frameworks, intelligent agents can engage in higher-order thinking, planning, and decision-making, thereby demonstrating higher levels of intelligence in complex cognitive tasks. Firstly, autobiographical memory is a crucial element at the L4 level. This means that intelligent agents can store and recall information about individual life experiences, forming an internal memory system regarding their own history and experiences. This allows intelligent agents to comprehensively and deeply understand their past, providing essential references for future decisions and planning.

Another important element is narrative self. This implies that intelligent agents can organize and comprehend their experiences in a narrative manner, constructing a coherent personal story. By incorporating individual experiences into a cohesive narrative framework, intelligent agents can better understand their values, beliefs, and goals, laying the foundation for more profound reflections. Lastly, self-identity is another key element at the L4 level. This indicates that intelligent agents can abstractly understand their own identity, including their individual status in society and culture, as well as their relationships with others. This abstract self-awareness enables intelligent agents to gain a deeper understanding of their existence and role, providing support for higher-level cognitive activities. The elements of autobiographical memory, narrative self, and self-identity collectively provide intelligent agents with a more profound and comprehensive self-awareness, laying the groundwork for them to demonstrate higher levels of intelligence in complex cognitive tasks.

\section{Realizing Self-based Cognitive Functions for AI}
Based on the multi-hierarchical framework of the Self, BriSe AI aims at the self-organized synergy of multiple cognitive functions and different learning strategies to achieve human-level cognitive abilities. Corresponding to the different levels of Self, this paper realizes multiple Self-based cognitive functions for AI, ranging from perception and learning, bodily self-perception to multi-modal conceptual learning, from bodily perception and decision-making to multi-agent collaboration, from self-perception to distinguishing self from others, empathizing and understanding self and others, and manifesting altruistic moral behavior in social interactions.

\subsection{Perception and Learning}
BriSe AI's integrative approach has significantly enhanced its ability to adeptly navigate and address complex challenges across various learning domains, including unsupervised learning~\cite{dong2023unsupervised}, supervised learning~\cite{zhao2023improving}, few-shot learning~\cite{li2022n}, continual learning~\cite{han2023enhancing}, transfer learning~\cite{he2023improving}, multi-modal concept learning, and evolutionary learning~\cite{shen2023brain}. These capabilities are crucial for tasks such as advanced visual classification and recognition, where unsupervised and supervised learning play key roles, and few-shot and continual learning strategies are vital in dynamic environments. Particularly, the principles of reinforcement learning have been effectively utilized in complex tasks like robot and drone control, demonstrating adaptability and decision-making prowess in real-world scenarios. Moreover, these achievements in learning and application have also supported the development of higher-level Self within BriSe AI, contributing to its evolving understanding of the Bodily Self, Autonomous Self, Social Self, and Conceptual Self.

\textbf{Multi-modal Concept Learning.} The perception and cognition of multisensory inputs facilitate the learning of concepts. The multi-modal concept learning is the foundation for a comprehensive understanding of the environment, oneself, and others. In the field of cognitive psychology, the mechanism of human concept learning is primarily composed of three subsystems: the multimodal experience system, the language-supported system, and the self-based semantic control system. By studying the mechanisms of human concept learning and the self-model, we developed a brain-inspired concept learning spiking neural network model. This model integrates specific sensory modalities such as vision, hearing, smell, taste, and touch, along with abstract textual information. Through similarity concept tests and concept classification tests, we can demonstrate that the concept representations generated by our model align closely with human cognition.
\begin{figure}[!htbp]
    \centering
    \includegraphics[width=0.9\linewidth]{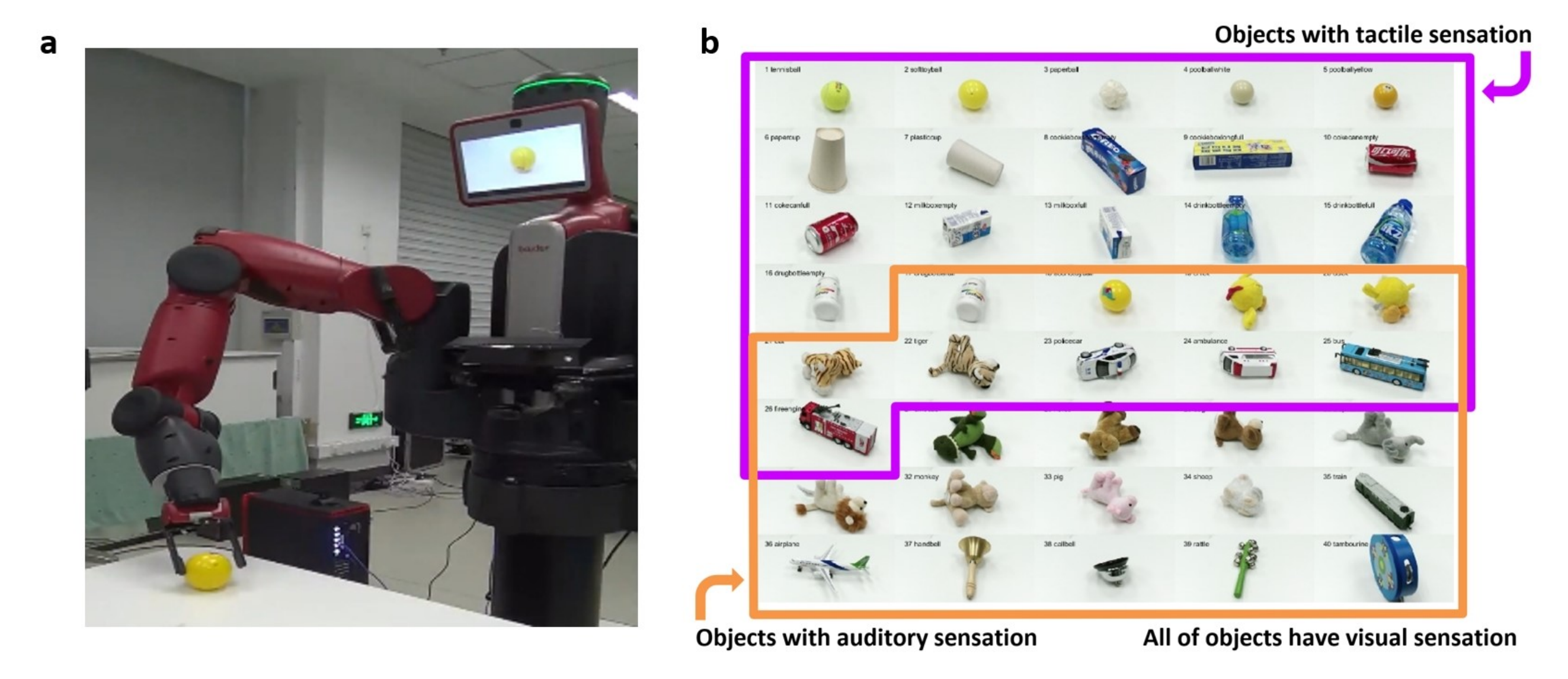}
    \caption{The Baxter robot for multi-modal concept learning. \textbf{a.} The Baxter robot. \textbf{b.} The 40-class multi-sensory dataset.}
    \label{fig:MultisensoryResults}
\end{figure}

The effectiveness of the model is validated through the robot Baxter, effectively bridging the gap between theoretical constructs and practical application.
The Baxter robot, from its own perspective, is capable of perceiving comprehensive multisensory information of objects, encompassing visual, auditory, and tactile modalities.
The visual information is captured by the camera in the chest of Baxter an additional Microsoft Kinect Sensor V2 camera for the higher resolution.
The auditory stimuli are recorded using Kinect Microphones. 
The tactile perceptions of object weight and compliance are implemented on our Baxter robot equipped with default 2-finger grippers.
As Figure~\ref{fig:MultisensoryResults} shows, in our multisensory classification task involving 40 distinct objects conducted on the Baxter robot, we achieved an accuracy rate of 94\%.

\subsection{Bodily Self} 

The acquisition of the Bodily Self is crucial for intelligent agents, as it enables the autonomous establishment of sensorimotor associations and the learning of distinctive features. Through Bodily Self, intelligent agents can enhance their cognitive abilities by autonomously constructing associations between movement and vision, and acquiring external appearance characteristics. This capability is instrumental in elevating the intelligence level of intelligent agents, aiding them in distinguishing themselves from others. Furthermore, it lays the foundation for autonomous exploration and social interactions, allowing intelligent agents to engage effectively in interactions within the social environment.

\textbf{Bodily Perception.} Bodily perception is one of the core elements of the bodily self. By integrating perceptual information from Level 0 Perception and Learning, intelligent agents engage in multimodal information integration, achieve motor-visual associative learning, construct a body model, and provide the foundation for high-level Self. A central question in bodily perception is how body illusions, such as the rubber hand illusion~\cite{Botvinick1998} occur. The rubber hand illusion refers to the phenomenon where subjects, when receiving tactile and visual stimuli simultaneously on a rubber hand and their unseen real hand, experience an illusion, perceiving the rubber hand as if it were their own.

 \begin{figure}[htbp]
\centering
\includegraphics[scale=0.39]{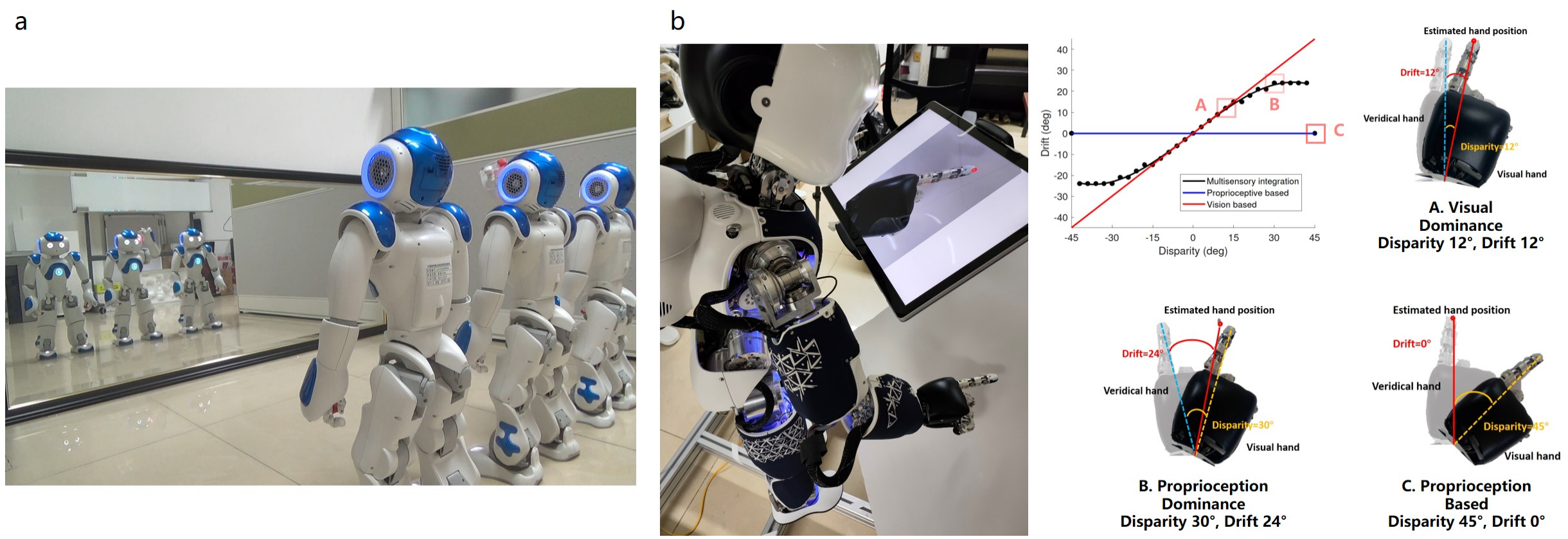}
\caption{Multi-robot mirror self-recognition experiment and rubber hand illusion on robot for self-perception.}
\label{RHI}
\end{figure}

We draw insights from biological studies on monkeys and humans and propose a brain-inspired bodily self-perception model~\cite{ZhaoPatterns2023}. This model understands and recognizes one's own body through the integration and association of multiple somatosensory perceptions and nonproprioception. It enables the robot to pass the mirror test~\cite{ZengCognitiveComputation2018} and induce the rubber hand illusion~\cite{ZhaoPatterns2023}. Mirror test is a classic experiment to determine whether animal species have self-awareness. The result of multi-robot mirror self-recognition experiment is shown in Figure ~\ref{RHI}a. Three robots with identical appearance move randomly in front of the mirror at the same time. The robot can identify which mirror belongs to it by comparing the trajectory predicted according to its own movement with the trajectory actually detected by vision. For example, when the robot is finished moving, the laser light is hit on the left side of the head of the middle robot, and the robot will touch the corresponding position. The robot obtained experimental results similar to those of monkeys in the rubber hand illusion experiment (as shown in Figure ~\ref{RHI}b). When the visual deflection angle is small, the robot's behavior decision mainly relies on visual information, and the proprioceptive drift increases with the increase of the visual deflection angle. When the visual deflection angle is medium, the behavior decision of the robot mainly relies on proprioceptive information. As the visual deflection angle increases, the proprioceptive drift increases slowly and tends to be flat. When the visual deflection angle is large, the behavior decision of the robot completely relies on proprioceptive information, and the proprioceptive drift is zero. The robot does not consider the hand in the field of view as its own.

\begin{figure}[htbp]
\centering
\includegraphics[scale=0.15]{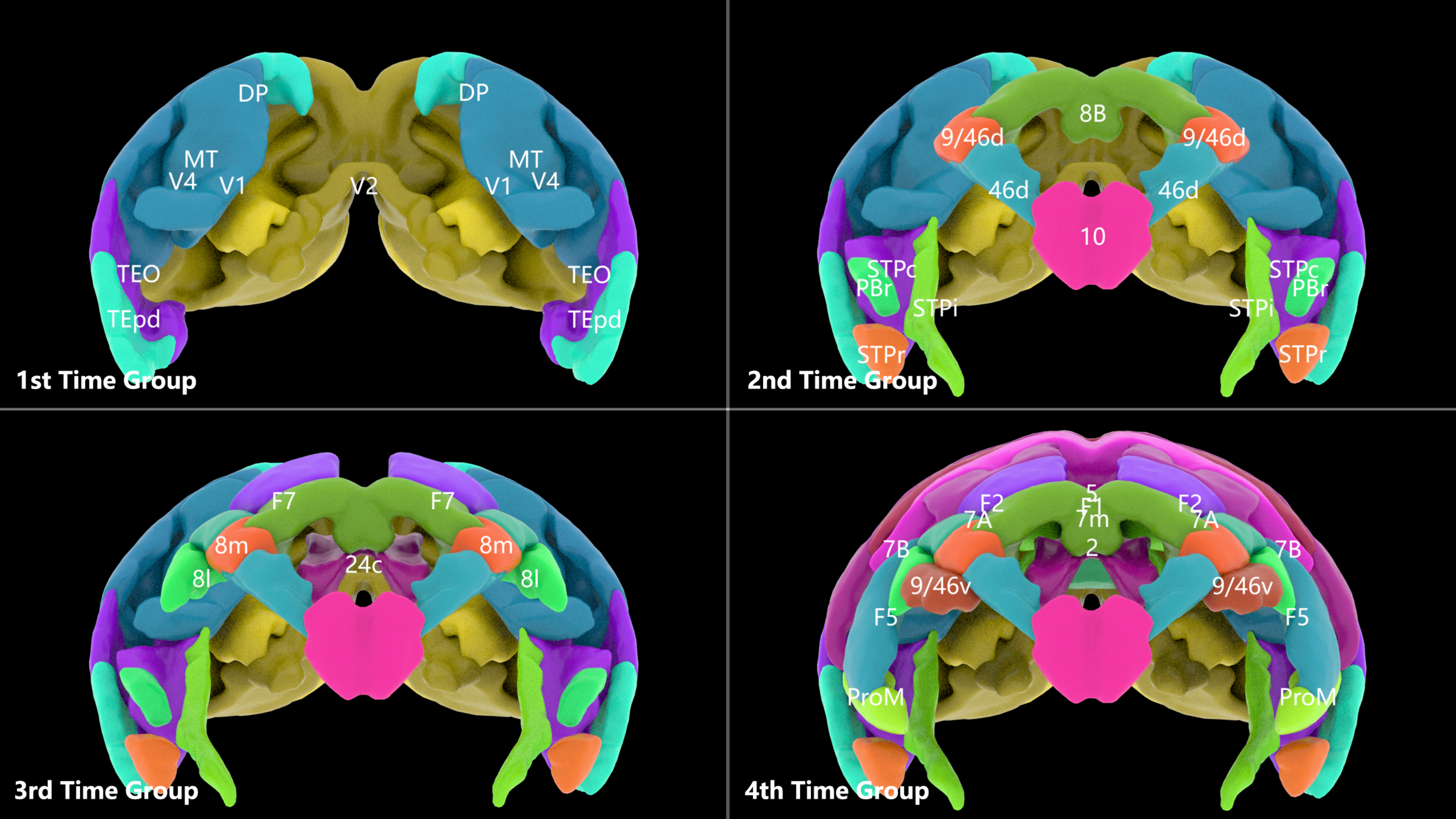}
\caption{Sagittal plane of the propagation path of rhythm diffusion.}
\label{bodily_3}
\end{figure}

The medial prefrontal cortex (MPFC) and its subregions play a crucial role in self-referential processing~\cite{d2007distinct}. We explored the anesthesia-induced loss of consciousness on the perception of bodily self based on a primate corticothalamic network model encompassing 29 widely distributed cortical areas cortex (including the medial prefrontal) and the thalamus~\cite{zhang2023hierarchical}. In recent years, anesthetics have emerged as novel instruments for elucidating the neural foundations of consciousness, enabling the precise and controlled simulation of both the loss and regain of consciousness. Propofol, in particular, has gained widespread use as a general anesthetic~\cite{alkire2008consciousness}. We introduces a brain simulation model including multiple brain regions to explore the neural activity during the propofol-induced loss of consciousness. Our model presents distinct advantages over conventional clinical observation data in that it offers quantifiable and controllable concentrations of propofol, along with high temporal resolution. During general anesthesia, bodily perception and consciousness diminish over time, correlating with the progressive involvement of different brain areas in rhythm propagation. This spatial distribution is distinct and systematic (as shown in Figure~\ref{bodily_3}), with a noticeable transition of rhythmic activity from the occipital lobe to the prefrontal and then to the frontal lobe across the first three-time groups. Ultimately, a complete cortical diffusion occurs. This model-generated process aligns with the clinically observed phenomenon of  $\alpha$-rhythm anteriorization~\cite{vijayan2013thalamocortical}.

\subsection{Autonomous Self}
For the Autonomous Self of BriSe AI, we realized classical conditioning and decision-making capabilities through autonomous exploration and interaction with the environment, associative learning, and trial-and-error learning. During this process, the agent continuously accumulates self-experience and understands the consequences of its own behavior, thus supporting a higher level of social self.

\textbf{Classical Conditioning.}
Classical conditioning is one of the most fundamental and crucial learning mechanisms in the Autonomous Self. It aids organisms in adapting to the complex and dynamic natural environment, allowing them to anticipate and adapt to environmental changes predictably, thereby enhancing their survival capabilities. One of the most renowned experiments in classical conditioning is conducted by Pavlov~\cite{Pavlov1927}. In this experiment, when dogs are presented with food (unconditioned stimulus, $US$), they exhibit salivation (unconditioned response, $UR$). In Pavlov's research, a sound (conditioned stimulus, $CS$) is introduced before each instance of food presentation. After numerous trials, the dogs start salivating upon hearing the tone (conditioned response, $CR$). 

Based on existing research findings in disciplines such as biology and neuroscience in the field of classical conditioning, we proposed a brain-inspired classical conditioning model~\cite{zhaoiScience}. This model integrates the biological research findings that have reached consensus in the field of classical conditioning into a spiking neural network. This model revolves around the interpositus nucleus (IPN), pontine nuclei (PN), granule cell (GC), Purkinje cell (PU), parallel fiber, climbing fiber, and inferior olive. It is designed to create pathways for conditioned reflexes and unconditioned reflexes. The synaptic plasticity mechanisms based on associative learning occur between PN and IPN, as well as between GC and PU. Compared to other computational models, the brain-inspired classical conditioning model can reproduce up to 15 classic conditioned phenomena and provide reasonable explanations from a computational perspective, which helps to reveal the biological mechanism by which organisms establish classical conditioning. In addition, the model can be deployed on robots, enabling them to establish classical conditioning similar to organisms. After verification, it has been demonstrated that this model endows robots with the capability of speed generalization. The speed generalization ability refers to the ability of robots to complete navigation tasks through conditional reflection at lower speeds and to complete navigation tasks without training at higher speeds. In a simulation environment with precise and controllable speed, the robot can adapt up to 3.5 times the speed and complete navigation tasks.

\textbf{Decision Making.}
Decision-making refers to the interaction of an agent with its ontologically perceived environment, autonomously adapting its strategy to accomplish a specific task based on environmental feedback. As a fundamental ability of the autonomous self, decision-making utilizes perception and learning at level 0 to understand the perceived environment. It leverages the bodily Self to provide a perception of the first perspective and the body's interaction with the environment. The rules acquired by the agent serve as its own experiences, enabling inference and induction of mental states. Self-experience is a key factor in distinguishing oneself from others and forms the foundation of social self and conceptual self.

\begin{figure}[!htbp]
\centering
\includegraphics[scale=0.4]{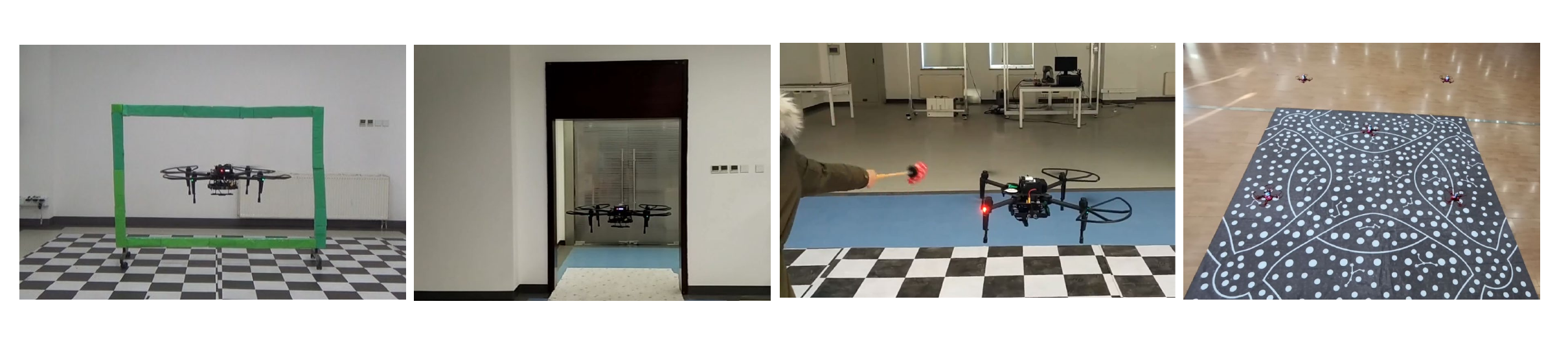}
\caption{Multiple decision-making tasks.}
\label{dm}
\end{figure}

To realize decision-making function, we designed the multi-brain region coordinated spiking neural network to model the decision-making neural circuit of the mammalian brain (prefrontal cortex-basal ganglia-thalamus-premotor cortex circuit)~\cite{zhao2018braina}. 
The excitatory and inhibitory connections between multiple brain regions are optimized by biologically plausible synaptic plasticity, where dopamine-modulated synaptic plasticity plays a key role in achieving trial-and-error learning. The decision-making process comprises subsystems including the agent's perception of the environment, winner-takes-all action selection, environmental feedback, strategy adjustment, and autonomous exploration. It collaboratively incorporates reinforcement learning strategies to acquire skills and accomplish tasks. As shown in Figure~\ref{dm}, the proposed model supports agents playing games~\cite{sun2022solving}, UAV flying through doors and windows~\cite{zhao2018brain}, obstacle avoidance in real-world scenarios~\cite{zhao2018braina}, and the safe flight of multiple drones within a limited area~\cite{zhao2022nature}.

\subsection{Social Self}
In terms of the social self, we realized the self-others distinction, perspective transformation based on self-experience achieved in Autonomous Self. Further, we separately constructed the theory of mind (ToM) and affective empathy SNNs to infer others' emotions, behaviors, goals, etc., aiding in better cooperation with others. Agents with Social Self show preliminary moral intuitions, such as altruistic rescue and helping others avoid safety risks.

\textbf{Theory of Mind/Cognitive Empathy.} Theory of Mind or Cognitive Empathy refers to a higher cognitive ability of individuals to understand their own and others' mental states, including intentions, expectations, thoughts, and beliefs, and use this information to predict and explain the behavior of others ~\cite{Premack1978}. One of the key milestones in the development of ToM is the acquisition of the ability to attribute false beliefs, recognizing that others may have different beliefs about the world ~\cite{Wimmer1983}. This ability requires the agent to first distinguish itself from others~\cite{ZengCogCom2018}, and based on this, predict the results of others' perception, and infer their beliefs and behaviors based on their own experience. This process closely relies on the Self's Level 0 and Level 1 to understand one's own perception, and the Level 2 Autonomous Self to accumulate self-experiences to distinguish self from others.

\begin{figure}[!htbp]
\centering
\includegraphics[scale=0.45]{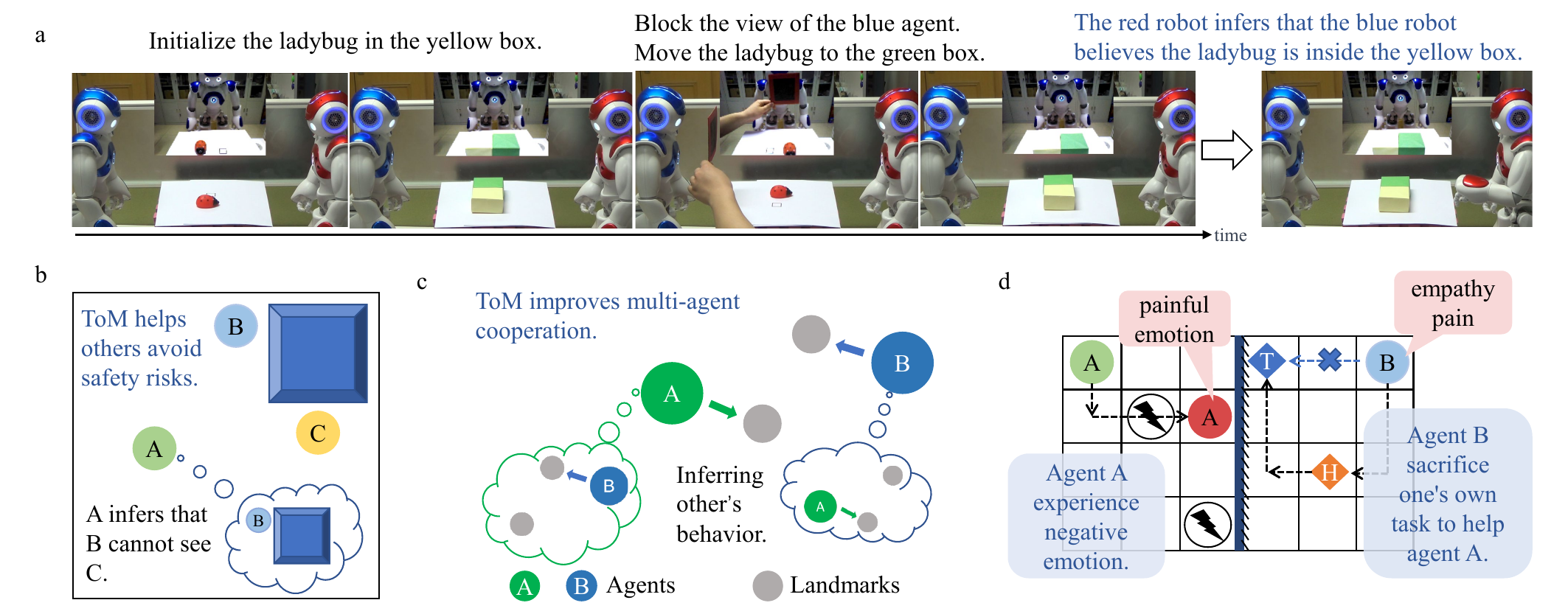}
\caption{The cognitive applications of Social Self. \textbf{a.} ToM enables robot to pass the false belief tasks. \textbf{b.} ToM helps others avoid safty risks. \textbf{c.} ToM improves multi-agent cooperation. \textbf{d.} Affective empathy empowers altruistic rescue.}
\label{tom}
\end{figure}

Drawing on research findings from cognitive psychology, neuroscience, and neuroimaging in the field of ToM, we integrate the multi-scale neural foundations of the theory of mind, including brain regions, brain functions, and neural circuits. We propose and construct a brain-inspired theory of mind SNN model~\cite {ZengFront2020}. The model allows robots to learn from their own experiences and use them to infer the beliefs of others, as well as predict their behavior based on their beliefs. This enables robots to acquire the ability to think critically through false belief tasks, as shown in Figure~\ref{tom}a.

In addition to conducting robotics experiments, we designed a virtual environment with perspective occlusion inspired by false belief tasks~\cite{baron1985does}. From Figure~\ref{tom}b, agents with obstructed vision are unable to avoid unknown safety risks while moving. However, other agents with ToM model are able to infer dangers they cannot see through perspective transformation, thus helping them to avoid danger. Besides, in multi-agent cooperative tasks, agents with ToM model are able to infer others' strategies based on self-experience or directly others-modelling. The inferred behavior of others influences one's own decisions to improve the efficiency of collaboration between agents (Figure~\ref{tom}c).

\textbf{Affective Empathy.} Affective empathy is the ability to perceive and understand the feelings or emotional states of others and is the motivation for altruistic behaviors such as comforting and rescuing~\cite{ref1}. It reflects the Social Self to a certain extent. The mirror neuron system~\cite{ref2}, as the core of affective empathy, empathizes the others' emotions by linking others emotional overt actions with their own emotions. The prerequisite for empathy is the self-emotional experience acquired at Level 2 Autonomous Self, which is the correlation between one's own experienced emotions and overt actions. Therefore, seeing similar others' overt actions triggers one's own previous emotional response.

Our proposed BriSe AI integrated a brain-inspired affective empathy computational model based on the mirror neuron system, empowering agents to empathize with others' emotions based on their own emotional experiences and to distinguish between self and others~\cite{ref3}. As shown in Figure~\ref{tom}d, we designed an affective empathy task where agents first experience the generation of their own emotions and emotional overt actions in an environment. Subsequently, they use their own experiences to empathize with their peers' emotions and carry out rescue behaviors. The experimental results show that when perceiving the peer's emotional overt action, it activates the same mirror neurons and emotional neurons as when the agent experiences its own emotion, realizing the sharing of emotion. Empathy for others' negative emotions becomes an intrinsic drive for altruism, motivating the agent to sacrifice one's own task to help others avoid negative emotions.

\section{Mutual Promotion between Learning and Self}

The core mechanism of BriSe AI is the mutual promotion between learning and Self, that is, self-based learning strategies support the acquisition of different levels of Self in a self-organized manner, and the Self in turn modulates and empowers the learning strategies with deeper understanding ability. The positive mutual promotion and support between Self and learning are the driving force empowering BriSe AI towards Artificial General Intelligence.

In the presence of Self, learning is not just the acquisition of information in the traditional sense, but a conscious understanding rooted in Self. 
For example, for the self-perception and ToM in BriSe AI, the process of distinguishing between self and others is not a simple classification problem, but rather an essential one based on the understanding of self to dig out the difference between self and others; Empathizing with others is not a direct recognition of others' emotion, but involves using one's own experience, scheduling one's own emotions, overt actions to deeply mirror and understand the emotions of others. Therefore, the Self has a positive impact on many aspects of the learning, breaking through the information processing of traditional learning methods to realize the true meaning of understanding oneself, the environment and others.

These self-based learning strategies support the realization of different levels of Self through self-organized coordination. For example, self-perception involves motor-visual associative learning,  as well as multimodal integration of vision and proprioception. Self-acquisition supported by learning strategies further modulates and monitors the underlying learning strategies, which in turn imparts new meaning to learning and deepens the understanding of intelligence. This reciprocal self-modulated learning infuses vitality and adaptability into the entire self-framework, enabling it to better understand and adapt to complex environments.

In this section, we first provide a comprehensive introduction of the fundamental learning strategies involved in BriSe AI. Next, we describe how these self-based learning strategies collaborate with self-related neural mechanisms to help realize different levels of Self. Finally, we describe how the Self in turn modulates various learning, forming a mutual promotion between learning and the Self.

\subsection{Fundamental Learning Strategies} 

In BriSe AI, a variety of fundamental learning strategies are employed to support the realization different levels of Self. These learning strategies draw inspiration from the biological brain's structure, mechanisms, and information processing. Incorporating multi-scale biological plasticities, BriSe AI integrates ten categories of SNN-based learning strategies, ranging from basic unsupervised learning, supervised learning, reinforcement learning, and associative learning, to few-shot learning, multi-modal concept learning, continual learning, transfer learning, and evolutionary learning. 

\textbf{Unsupervised Learning.} 
BriSe AI-integrated unsupervised learning is based on Spike-Timing-Dependent Plasticity (STDP), and introduces adaptive synaptic filters for regulating regulate input currents, adaptive lateral inhibitory connections to ensure diverse feature extraction, and adaptive threshold balancing for maintaining the input-output balance across neural layers~\cite{dong2023unsupervised}. Together, these adaptation mechanisms improve the network's information processing, learning efficiency, and feature representation capabilities.

\textbf{Supervised Learning.}
Traditional SNN training algorithms mainly focus on the guiding role of label information in predictions, often neglecting the variability in the network's output distribution across different time steps. This variability can lead to conflicts in the direction of optimization during the training process, thereby hindering effective optimization of the network. Similarly, during the testing process, differences in distribution between time steps can affect the network's performance due to inaccurate output results. Our supervised learning algorithm~\cite{zhao2023improving} enhances the consistency across different time steps, thereby improving the stability of SNN training and significantly enhancing performance under low-latency conditions. This algorithm combines traditional cross-entropy loss with a novel enhanced temporal consistency loss, the latter reducing differences between time steps through Kullback-Leibler divergence.

\textbf{Reinforcement Learning.}
We adopt a layered membrane potential regularization method, and construct an efficient brain-inspired spiking neural network model for deep reinforcement learning with efficient spike information transmission~\cite{sun2022solving}. Inspired by the brain's state value distribution decision-making mechanism, we employ neuronal population coding to enhance the representational capability of spike sequence information, creating Gaussian neuron populations for the dimensional expansion and separable representation of binary spike sequences corresponding to continuous quantile values. Drawing inspiration from the structure of brain pyramidal neurons, we use multi-compartment neural networks to address the fusion of decision-making information from diverse sources ~\cite{sun2023multi}. The population neuron coding approach and multi-compartment neuron information integration structure are more effective at handling multi-source decision information than the commonly used point neuron models and membrane potential computation methods in spiking neural networks.

\textbf{Associative Learning.} Motor-visual associative learning is the process of establishing connections between self-generated movements and visual feedback during the learning of motion. The somatosensory pathway and the dorsal visual pathway are the main pathways involved in this process. If signals from these pathways occur simultaneously within the same time window, new connections may form or existing connections may be strengthened to represent their association. Appearance learning is the process of learning the appearance of body parts when the expected visual feedback is consistent with the movement. The proprioceptive pathway and the ventral visual pathway are the main pathways involved in this process. Through motor-visual associative learning, robots can associate expected visual feedback with their self-generated movements. When these self-generated movements are executed again, real visual feedback is received. If this real visual feedback is consistent with the expected feedback, relevant brain areas are activated, connections are established through synaptic plasticity mechanisms, and the robot learns the appearance of body parts.

\textbf{Few-shot Learning.} We develop a neuromorphic dataset that reflects the complexities of human visual perception and cognition, N-Omniglot~\cite{li2022n}. The dataset comprises an extensive array of handwritten characters, each category represented by a minimal number of examples. In addition, we have developed a variety of few-shot learning methods, including improved nearest neighbor, convolutional network, SiameseNet, and meta-learning algorithm, demonstrate the versatility and efficacy of N-Omniglot. These methods are carefully tailored to exploit the unique characteristics of the N-Omniglot dataset, exemplify BriSe AI's multifaceted approach to tackling complex AI challenges.

\textbf{Multi-modal Concept Learning.} We develop a computational model for concept learning that accurately emulates the architectural paradigm of human cognition\cite{wywmultisensory}\cite{wywtri}. 
The tripartite structure of the computational model - consisting of a multisensory information processing module, a text-derived information processing module, and a self-based semantic control module - mirrors the three networks of human cognition: a multimodal experiential system, a language-based semantic system, and a semantic control system.
The main idea of our human-like concept learning is that despite the myriad of external inputs encountered by the human brain during the learning of new concepts, these inputs are all processed as unified signal transmissions within the brain. 
We convert the two types of concept representations into discrete neural stimuli, transforming the original representation into a spike train, thereby standardizing the data representation and simulating the initial process of concept acquisition in the human brain. 
What's more, we innovatively leverage the properties of spiking neural networks\cite{wywstat}. 
The information in the temporal domain is harmonized by controlling the duration T of the performance of neurons simulated by external information, and the information from the two representations is integrated via sliding coordination. 
The final human-like concept representation is derived once the concept representation has been coordinated both spatially and temporally.

\textbf{Continual Learning.} The human brain possesses the ability to learn multiple cognitive tasks sequentially and its structural developmental mechanisms allow the nervous system to dynamically expand and contract to flexibly invoke task-specific neural circuits. Motivated by this, a Dynamic Structure Development of Spiking Neural Networks (DSD-SNN) is proposed for efficient and adaptive continual learning~\cite{han2023enhancing}. When a new task is coming, the DSD-SNN model first randomly grows a part of neurons forming a new pathway to learn new knowledge and receive old knowledge that can be reused. During training for new tasks, the DSD-SNN gradually prune the redundant neurons that are continuously inactive for several epochs. Dynamic neuronal expansion and pruning increase memory capacity and reduce energy consumption, and the overlapping shared structure helps to quickly leverage all acquired knowledge to new tasks.

In addition to a continual learning approach based on dynamic development, we also implement a continual learning model based on local synapse importance evaluation and sleep-induced memory consolidation. This model has two phases: waking and sleep. In the waking stage, the model realizes the important weights of different tasks through the local computation rule, the Hebbian algorithm. The important weights are constrained to not change largely so as to restrain the learning process of new tasks successively, reducing the interference of old knowledge. During sleep, the homeostasis realized by a global adjustment of the network activities enhances the working memory and neural replay by the reorganized connection between the working and long-term memories, to further consolidate the long-term memory.

\textbf{Transfer Learning.} Transfer learning enables spiking neural networks to adapt knowledge from static image data and then effectively transfer and apply this knowledge to event-driven data processing. This self-adaptive ability enhances the network's ability to generalize when processing event data and improves the network's flexibility and adaptability across multiple data types. In detail, we propose a knowledge transfer method that combines domain alignment loss and spatio-temporal regularization. This method learns domain-invariant spatial features by reducing the marginal distribution distance between static images and event data. Meanwhile, spatio-temporal regularisation captures temporal attributes in the data by providing dynamically learnable coefficients for the domain alignment loss, which ensures a specific weight assignment to the data features at each time step.
Our method is represented as a loss function that constrains and guides the network to learn. The knowledge transfer loss can be expressed as:
\begin{align}
  \small
  \mathcal{L}_{kt} &= 1 - \frac{1}{T}\sum_{t=1}^T \sigma(\eta_t)\mathop{CKA^{\prime}}\limits_{y_i = y_j, y \in \mathcal{Y}}\left(g\left(\mathbf{x}_s^i, t\right), g\left(\mathbf{x}_t^j, t\right)\right) + \frac{1}{T}\sum_{t=1}^T(1-\sigma(\eta_t))\ell_{cls-e}
  \label{Eq15}
\end{align}
where $\eta_t$ denotes the learnable coefficient at time step $t$ and $\sigma$ represents the sigmoid function. 
$\mathop{CKA^{\prime}}$ represents the computation of the kernel function of the vectors followed by the computation of central kernel alignment (CKA) \cite{kornblith2019similarity}. $g\left(\mathbf{x}_s^i, t\right)$ and $g\left(\mathbf{x}_t^j, t\right)$ denote the features of the data in the source and target domains respectively after the encoder $g$. $\ell_{cls-e}$
is classification loss of event data.

\textbf{Evolutionary Learning.}
For evolutionary learning, the NeuEvo~\cite{shen2023brain} framework introduces an integrated approach for evolving the neural circuitry of SNNs, inspired by the dynamic and adaptive nature of biological neural systems. This integrated approach is not solely focused on evolutionary computation, but rather encompasses a comprehensive strategy that blends local synaptic dynamics with global network optimization. The framework leverages a synergistic process for synaptic weight adaptation. This process combines local synaptic activity, governed by the STDP rule, with global error feedback for network-wide optimization. The synaptic weight (\( w \)) at any given neuron is updated based on the following formulation:

\begin{equation}
    w_{t+1} = w_t + \eta \cdot \Delta w_{local} + \beta \cdot \Delta w_{global}
\end{equation}

Here, \( w_t \) represents the synaptic weight at time \( t \), \( \eta \) and \( \beta \) are the learning rates for local and global updates, respectively, and \( \Delta w_{local} \) and \( \Delta w_{global} \) are the weight changes due to local and global learning processes.

The structural configuration of neural circuits in the NeuEvo framework is dynamically adjusted based on the interplay of local and global learning signals. Initially, a diverse and dense connection pattern is established, simulating the potential synaptic pathways in the network. These connections are then selectively strengthened or weakened based on their contribution to the network's performance, as well as their efficiency and reliability in information processing. The adjustment of these connections is driven by a balance between reinforcing beneficial pathways and pruning less effective ones, thereby optimizing the network's architecture for specific tasks.

In addition, the ELSM~\cite{pan2024emergence} and MSE-NAS~\cite{pan2023multi} specially design adaptive structure evolution for recurrent and deep spiking neural networks, respectively. Through exploring the small-world characteristics that exist in the human brain, ELSM guides the evolution of network topology from both static and dynamic perspectives. MSE-NAS evolves microscopic neuron computations, mesoscopic microcircuit combinations, and macroscopic global brain area connectivity, supplemented by customised brain-inspired evaluation function, encoding scheme, and genetic operations. Adaptive LSM~\cite{pan2023adaptive} deeply draws inspiration from brain evolution mechanisms, integrating structural evolution with multi-scale biological plasticity learning rules.

\subsection{Self-organized Learning supported Self Implementation}
Different learning strategies organized in a self-organized manner, support the realization of different levels of Self, ranging from perception and learning to bodily Self, autonomous Self, social Self, and conceptual Self. For example, embodied agent integrates few-shot learning, unsupervised and supervised learning, associative learning, multimodal concept learning, transfer learning during perceiving the environment. Reinforcement learning, online learning, and continual learning engage in agent-environment interactions, and supervised learning, reinforcement learning, and evolutionary learning are required in multi-agent cooperation. These learning strategies also incorporate self-related neural mechanisms and neural circuits in the brain to realize self-enabled cognitive functions.
For instance, bodily self-perception establishes the relationship between self-visual and motor through associative learning, and combines the brain's prediction and expectation mechanisms to distinguish oneself from the environment and others. Autonomous decision-making utilizes dopamine-regulated neural circuits to achieve reinforcement learning in interactions with the environment, continuously accumulating self-experience. ToM utilizes one's own experience to make predictions about others, while incorporating the human brain's self-perspective inhibition mechanism to understand others in their situation. Affective empathy links overt actions and emotions through the mirror neuron system, thereby triggering one's own emotions through the visual perception of others' overt actions.

\textbf{Self-world distinction for bodily self-perception.} Motor-visual associative learning enables robots to understand the visual feedback caused by their own movements. If a robot detects motion within its field of view that is inconsistent with the expected visual feedback, it can determine that the motion is not its own. Conversely, if the motion detected matches the expected visual feedback, the robot can conclude that the motion is self-generated, thereby distinguishing between self and the external world.

\textbf{Dopamine regulation for decision-making.}
We constructed brain-inspired decision-making spiking neural network models based on multiple biological evidence, including the relevant brain regions involved in decision-making and their collaborative mechanisms. Spiking neurons form sub-regions that connect with each other to create mutually coordinated subsystems. The connections between spiking neurons are optimized using biological plasticity rules, such as local STDP synaptic plasticity and dopamine-regulated synaptic plasticity.

As a neurotransmitter, dopamine plays a critical role in emotions, memory, action selection, motivation, attention, reward-driven behavior, and motor control~\cite{wise2004dopamine}, and thus provides constructive inspiration for the development of the autonomous self. The neuromodulatory role of the dopamine system is also responsible for trial-and-error learning in decision-making, whereby the reward prediction error signal carried by dopamine directly modulates the connection weights from the prefrontal cortex to the striatum. 
Specifically, global long-term neuromodulation enables the responsiveness of local synaptic plasticity, yielding dopamine-regulated local synaptic plasticity. In this paper, We adopt the reward-modulated STDP~\cite{Eugene2007} that is capable of solving the problem of credit assignment.

\textbf{Self-perspective inhibition for ToM.} 
Inhibitory control mechanisms are considered crucial for the process of ToM, as they enable the output of information perceived from others' perspectives and others' beliefs, facilitating cognitive functions like understanding others' intentions, inferring others' beliefs, and making behavioral decisions based on others' beliefs. The inhibitory control mechanism includes inner inhibitory control and outside inhibitory control within the brain. When conflicting information between self and others arises, the inhibitory control within brain regions fails to inhibit self-relevant information due to the dominance of self-information. In such cases, other brain regions (such as IFG) are activated through outside inhibitory control to inhibit self-dominant information. 

Utilizing the inhibitory control mechanism of a single neuron as exemplified in our false belief task~\cite{ZengFront2020}, the Inferior Parietal Lobule (IPL) or ventromedial Prefrontal Cortex (vmPFC) receive currents related to both self-relevant and other-relevant stimuli. The currents feeding into inhibitory and temporary neurons vary depending on whether the inference involves others' beliefs (encompassing self-perspective and self-belief inhibition) or self-beliefs (including other-perspective and other-belief inhibition). During self-perspective and self-belief inhibition phases, the currents to inhibitory neurons align with other-related stimuli, whereas currents to temporary neurons correspond to self-related stimuli. Consequently, other-related stimuli are initially processed, followed by self-related stimuli. Conversely, in the other-perspective and other-belief inhibition phases, inhibitory neurons exhibit intense activity, capable of completely suppressing other-relevant stimuli, while inputs to temporary neurons mirror other-related stimuli. Thus, self-related stimuli precede processing, followed by other-related stimuli.
 
 Throughout the belief inference process, the spiking signals depicting self-perspective information are suppressed by inhibitory control mechanism. The spiking signals representing others' perspective perception, obtained through perspective transformation, are then processed in the model. This allows for the inference of others' beliefs using information from others' perspective perceptions and self-experience. In the process of behavioral decision-making, the spiking signals representing self-beliefs are suppressed by the inhibitory control mechanism, and the spiking signals representing others' beliefs are output as the basis for behavioral decision-making.

\textbf{Self-experience inferring others.} 
In the context of ToM and affective empathy, self-experience aids in inferring others' mental states and empathizing with others' emotions. This concept is also referred to as 'self-projection'~\cite{buckner2007self}, involves using personal memories to understand others' perspectives~\cite{moreau2013using}. Self-experience is derived from the interaction between the decision model and the environment. When constructing the ToM model, we incorporated a buffer for storing self-experience and a sub-module for inferring others~\cite{zhao2023brain}. A SNN-based sub-module was developed to simulate the mapping from self-experience learning to inferring others' behaviors. By using observations of others as input, our model can predict their behaviors.

\textbf{Mirror neuron system for affective empathy.} Psychological research suggests that observers can use the perception-action mechanism (PAM) to empathize with others~\cite{ref4}. Perceiving another person's emotional state activates the same emotional representation in the observer's brain, which is equivalent to the observer also experiencing the emotion, and the mirror neuron system (MNS) provides the biological basis for PAM~\cite{ref5,ref6}. The mirror neuron system contains mirror neurons and anti-mirror neurons. Mirror neurons enable individuals to empathize with others' emotions through their own emotional experiences, while anti-mirror neurons help distinguish the originator of the emotion, reflecting primary self-awareness~\cite{ref7}. Our proposed mirror neuron system-based affective empathy SNN contains three clusters of neuronal populations: the emotional cortex representing different emotional states, the motor cortex containing the mirror neuron system, and the perceptual cortex representing emotional overt actions. These modules are interconnected with the mirror neuron system as the core, and local synaptic plasticity is used to learn the connection strength, achieving empathy for others' emotions. Additionally, we replicate the emergence of anti-mirror neurons, achieving self-other differentiation.

\subsection{Self-modulated Learning}

An important characteristic of BriSe AI is that Self, implemented based on multiple learning strategies, in turn modulates learning to achieve conscious information understanding. As a result, self-based learning supports increasingly higher levels of Self, and the mutual reinforcement between Self and learning helps BriSe AI emerge as a human-level cognitive abilities. Specifically, learning without Self achieves information processing capability through biologically-inspired structures and plasticity, enabling unconscious learning and skill acquisition. However, knowledge acquired through underlying information associations does not equate to true understanding. Human-level intelligence is closely intertwined with the body. Cognitive processes and understanding in the brain are closely related to our bodily states and interactions with the environment. A conscious living brain is capable of scrutinizing and supervising the underlying learning process, autonomously activating neural circuits to discover unusual problems (e.g., information that contradicts common sense), and self-organizing to collaborate with the building blocks to deeply comprehend and infer implicit knowledge.

For example, in the case of robot multi-modal concept learning of BriSe AI, without Self, different modalities are overlaid without distinction in data-driven concept learning. However, the perception of an embodied robot is closely related to its self-modeling. The robot understands its own body through self-modeling, which forms the basis for perceiving the world and solving problems. This self-understanding determines which modalities the robot uses and how it combines multi-modal information with its own sensors to better understand the real world. Self-based robot multi-modal concept learning breaks away from data-driven learning to a full and comprehensive understanding of objects based on self-knowledge, driven intrinsically by oneself. Associative learning that supports self-perception also goes beyond establishing superficial associations of information, but delves into the understanding of the Self and defining itself. Only with such self-perception can an agent derive the ability to understand others in social interactions, such as ToM and emotional empathy. Unlike traditional reinforcement learning, which requires extensive interaction for data acquisition, a self-modulated agent is intrinsically motivated to consciously and actively explore novel and unusual environments, and continuously accumulate experience online while gaining in-depth understanding and reasoning. Overall, self-modulated learning is committed to realizing the leap from information processing to information understanding, thereby better supporting different levels of Self. The positive mutual enhancement between Self and learning makes BriSe AI more adaptive and capable of advanced intelligence.

\section{Discussion and Conclusion}

In this paper, we introduce BriSe AI, a Brain-inspired and Self-based Artificial Intelligence paradigm. We illustrate the crucial role that BriSe AI plays in realizing AGI and the future harmonious coexistence of humans and machines, distinguishing it from existing cognitive architectures and AI methodologies. BriSe AI is based on the hierarchical framework of Self, including Perception and Learning, Bodily Self, Autonomous Self, Social Self, and Conceptual self. BriSe AI is based on multi-scale biological plasticities, supporting more complex and advanced self-based cognitive capabilities within a unified framework by self-organized synergy of multiple cognitive functions and various learning strategies. Specifically, fundamental perception and learning combined with bodily self-perception assists the AI agent in understanding both the environment and itself. Active exploration and interaction with the environment enables the agent to acquire task execution, perceive its spatial and temporal position, and form self-experience as well as develop self-causal awareness and self-awareness. Crucially, BriSe AI focuses on the ability of Self in social interactions, such as distinguishing self from others, understanding self and others, seeing others as agents like you, and having empathy for others, which are key aspects for the future symbiotic society where human and AI coexist. For higher levels of Self, the concept self embodies an abstract understanding of self identity, values and goals.

BriSe AI is not limited to achieve different levels of Self based on multiple learning strategies, and the gradual development of higher-level Self from lower-level Self. More importantly, Self can, in turn, aid in better learning, achieving a leap from information processing to a deeper understanding of information. Consequently, a positive mutual promotion has formed between Self and learning, with self-based learning gradually enhancing self-capability, and Self in turn enhancing the efficiency and adaptability of learning. Furthermore, the interconnections among the multiple levels of Self allows higher-level Self to supervise and modulate lower-level Self, so that the low-level Self improves its ability while better serving the high-level Self. Overall, the positive promotion among multiple levels of Self, and between Self and learning, enhances BriSe AI's conscious understanding of information and flexible adaptation to complex environments, marking a step towards advanced Artificial General Intelligence.

In terms of implementation, BriSe AI employs a full spiking neural network modeling approach, that mimics the information processing in the brain. It is built upon the established BrainCog infrastructure, organically integrating the fundamental components and cognitive functions of BrainCog. This makes BriSe AI more biologically realistic and biologically plausible. The BriSe AI has realized the multi-modal environmental perception and concept learning, bodily self-recognition, autonomous self of decision-making and classical conditioning, and social self for cognitive empathy and affective empathy. To the best of our knowledge, this work is the first to introduce a comprehensive and clear self-framework into an AI cognitive architecture, and implement various self-related cognitive functions, preliminarily demonstrating the potential of an artificial general-purpose AI.

Overall, this paper aims to provide a new paradigm for current and future AI research, a roadmap for AI rooted in the Self, which aims to build biologically realistic AI in the real sense, and serve as a foundation for harmonious coexistence between human and AI. Looking ahead to the long-term future, BriSe AI should be an important foundation to realize real Artificial General Intelligence, and Superintelligence, emerging as a novel evolutionary becoming, and serving as a moral partner for human in the future's symbiotic society. 

\bibliography{brise}

\end{document}